\documentclass[letterpaper]{article} 
\usepackage{aaai2026}  
\usepackage{times}  
\usepackage{helvet}  
\usepackage{courier}  
\usepackage[hyphens]{url}  
\usepackage{graphicx} 
\usepackage{natbib}  
\usepackage{caption} 
\usepackage{algorithm}
\usepackage{algorithmic}

\usepackage{amsmath} 
\usepackage{multirow}
\usepackage{amsmath,amsfonts}
\usepackage{textcomp}
\usepackage{makecell}
\usepackage{colortbl}  
\usepackage{collcell}
\usepackage[table]{xcolor} 
\usepackage{threeparttable}
\usepackage{bbding}  

\usepackage{newfloat}
\usepackage{listings}
\DeclareCaptionStyle{ruled}{labelfont=normalfont,labelsep=colon,strut=off} 
\floatstyle{ruled}
\newfloat{listing}{tb}{lst}{}
\floatname{listing}{Listing}
\title{
On Modality Incomplete Infrared-Visible Object Detection: An Architecture Compatibility Perspective
}
\author{
    Shuo Yang,
    Yinghui	Xing,
    Shizhou	Zhang,
    Zhilong	Niu,
}
\affiliations{
    Northwestern Polytechnical University, China\\
    \{shawny2023,niuzhilong\}@mail.nwpu.edu.cn, \{xyh\_7491,szzhang\}@nwpu.edu.cn
}

\begin{document}

\maketitle


\begin{abstract}
Infrared and visible object detection (IVOD) is essential for numerous around-the-clock applications. 
Despite notable advancements, current IVOD models exhibit notable performance declines when confronted with incomplete modality data, particularly if the dominant modality is missing. 
In this paper, we take a thorough investigation on modality incomplete IVOD problem from an architecture compatibility perspective.
Specifically, we propose a plug-and-play Scarf Neck module for DETR variants, which introduces a modality-agnostic deformable attention mechanism to enable the IVOD detector to flexibly adapt to any single or double modalities during training and inference.
When training Scarf-DETR, we design a pseudo modality dropout strategy to fully utilize the multi-modality information, making the detector compatible and robust to both working modes of single and double modalities.
Moreover, we introduce a comprehensive benchmark for the modality-incomplete IVOD task aimed at thoroughly assessing situations where the absent modality is either dominant or secondary.
Our proposed Scarf-DETR not only performs excellently in missing modality scenarios but also achieves superior performances on the standard IVOD modality complete benchmarks.
Our code will be available at https://github.com/YinghuiXing/Scarf-DETR.
\end{abstract}



\section{Introduction}
\label{sec:intro}

Infrared-visible object detection (IVOD) plays a crucial role in multi-modal learning~\cite{xu2024cross,zhao2023tstr,wang2023sgfnet,wu2023vehicle,piekenbrinck2024rgb} because infrared images provide supplementary information in challenging environments such as poor lighting, fog, or situations where visible images are difficult to capture targets. Existing methods typically focus on designing specialized infrared-visible object detectors, which often necessitate complex fusion modules, customized loss functions, and tailored optimization strategies~\cite{li2024fd2,shin2024complementary,xing2024ms,dams24}.

Despite significant progress, these models usually require all modalities to be complete during training, limiting their application in real-world scenarios.
In practical applications, sensor faults, blurriness and overexposure in images may occur, leading to the missing of specific modality. 
Given that the architectures of traditional IVOD models are often tightly coupled with the complete input of the modality, when a certain modality is missing, such models will exhibit significant deterioration.


\begin{figure}[t]\footnotesize
  \centering
  \begin{minipage}[c]{0.45\textwidth}  
    \centering
     \includegraphics[width=1.0\linewidth]{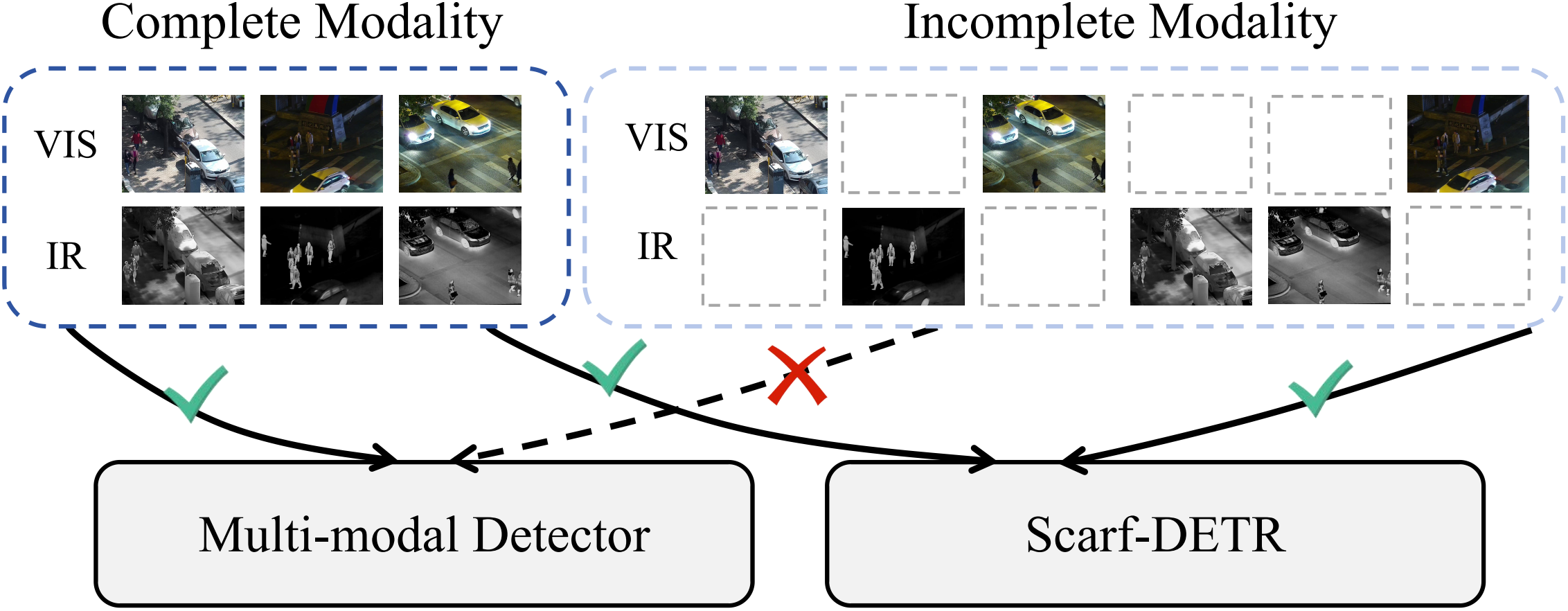}
  \end{minipage}
  \begin{minipage}[c]{0.45\textwidth} 
    \centering
    \setlength{\tabcolsep}{1.3mm} 

\begin{tabular}{ c| c | c c c} 
\hline
\multirow{2}{*}{\textbf{Modality}} &\multirow{2}{*}{\textbf{Model}}  & \multicolumn{3}{c}{\textbf{mAP} \(\uparrow\)} \\ 
\cline{3-5}
& & FLIR-aligned  &M\(^3\)FD &LLVIP \\  
\cline{1-5}
\multirow{2}{*}{\makecell[c]{VIS+IR}}
& DAMS-DETR  & 49.3           &52.9              &69.6 \\  
& Scarf-DETR & \textbf{49.6}  & \textbf{54.4}    &69.6  \\ 
\hline
\multirow{2}{*}{\makecell[c]{VIS}}
& DAMS-DETR  & 28.4          & 46.6            & 15.0         \\  
& Scarf-DETR & \textbf{37.8} & \textbf{51.4}   & \textbf{55.5}\\ 
\hline
\multirow{2}{*}{\makecell[c]{IR}}
& DAMS-DETR  & 47.0          & 26.9           & 69.3  \\  
& Scarf-DETR & \textbf{47.3} & \textbf{41.0}  & 69.3  \\ 
\hline
\end{tabular}
  \end{minipage}
  \caption{Illustration and necessity to deal with the modality-incomplete issue in IVOD. Available multi-modal detectors fail to detect targets with incomplete modality, while Scarf-DETR is compatible with complete and incomplete modality scenarios.}
  \label{fig:debut}        
\end{figure}

As shown in Figure~\ref{fig:debut}, previous state-of-the-art model DAMS-DETR achieves only 15\% mAP in the VIS-only scenario on the LLVIP dataset, which is 54.6\% decrease compared to the complete-modality setting.
We also observe that DAMS-DETR can achieve comparable results in the IR-only scenario compared to the ideal modality-complete setting (69.3 vs. 69.6).
Based on the observation, it can be clearly seen that the IR modality shows dominant while the visible modality shows secondary on the LLVIP dataset because it mainly consists of images captured at night or low-illuminance conditions.
Although we admit that whether the missing modality is dominant or secondary is a significant factor in the final performance, it is worth noting that our Scarf-DETR, which is designed to decouple double input modalities and the model architecture, can improve performance on secondary modality input scenarios from 15.0\% to 55.5\% in the LLVIP dataset when entering the visible modality only.
Consequently, having architectures that can seamlessly adjust to handle any individual or combined modalities during both training and inference is essential to ensure the robustness and dependability of IVOD detectors. 



In this paper, we devote our efforts to a comprehensive study on the modality incomplete IVOD problem from an architecture compatibility perspective, aiming to design a plug-and-play module which can flexibly adapt the detector to any single or double input modalities when equipped with existing detectors.
To this end, we develop a compatible neck
architecture alongside an effective training strategy to enhance performance and robustness.
Specifically, we introduce a versatile neck, termed as Scarf Neck, tailored for DETR-like detectors. 
This component incorporates a straightforward Modality-Agnostic Deformable Attention (MADA) mechanism, which is designed to efficiently integrate features from multiple modalities or to enhance features from a single modality when the other is absent.
In complete-modality scenarios, the method fully considers features from all modalities to dynamically select feature points in each modality and update the features of each modality separately. 
This approach, on the one hand, alleviates the prevalent spatial misalignment issue in multimodal images; on the other hand, it simultaneously achieves intra-modal feature enhancement and cross-modal feature fusion. 
In modality-missing scenarios, instead of simply duplicating the available modality to fit the subsequent model architecture, we align the model's data flow with an additional set of reference points to enhance the feature representations of the remaining modality.

To improve the robustness of the detector, we introduce a modality dropout mechanism to mimic the modality-missing scenarios during training. 
However, the naive modality dropout strategy reduces the diversity of training samples, thereby hindering model convergence. 
Therefore, we propose a pseudo modality dropout strategy that selectively disconnects modality connections for some sample pairs, enhancing both information utilization and performance in both complete and incomplete modality scenarios.

Moreover, to fully evaluate the performance of modality incomplete IVOD task, we construct a comprehensive benchmark based on existing three IVOD datasets, dubbed FLIR-MI, M$^3$FD-MI and LLVIP-MI.
While the visible and the IR modality images alone on each dataset can serve as significant benchmarks, we further randomly discard 30\%, 50\% and 70\% visible images and 70\%, 50\% and 30\% IR images of all the paired images respectively to constitute FLIR-MI, M$^3$FD-MI and LLVIP-MI with all single modality images to mimic the practical scenarios where the absent modality is either dominant or secondary. Furthermore, we conduct thorough experiments on the constructed three modality incomplete benchmarks with existing state-of-the-art IVOD detectors.


Integrating the Scarf Neck and pseudo-modality dropout strategy, our proposed Scarf-DETR achieves significant performance gains in modality-incomplete scenarios including both the VIS/IR only extreme cases and the constructed three modality incomplete settings.
Notably, our method also achieves superior or competitive performance compared with existing state-of-the-art IVOD detectors on three IVOD datasets. 
Figure~\ref{fig:debut} demonstrates that Scarf-DETR substantially outperforms DAMS-DETR~\cite{dams24} across all testing settings, with a remarkable 40.5\% mAP improvement in the VIS-only scenario on the LLVIP dataset. 
The main contributions of this paper are as follows:

\begin{itemize}
    \item We aim to conduct a comprehensive investigation into the modality-incomplete IVOD problem, with key research efforts including exploring this issue from the perspective of architecture compatibility and establishing a comprehensive evaluation benchmark.
    
 
    \item We develop a Scarf Neck module which utilizes modality-agnostic deformable attention to accommodate the input of any single or double modalities. Combined with our pseudo-modality dropout training strategy, Scarf Neck can serve as a plug and play module with DETR-series detectors. 
    
   
    \item Our proposed approach yields significant performance gains in modality-incomplete scenarios across the three constructed modality-incomplete benchmarks, namely FLIR-MI, M$^3$FD-MI and LLVIP-MI.
    
\end{itemize}

\begin{figure*}[ht]
  \centering
   \includegraphics[width=1.0\linewidth]{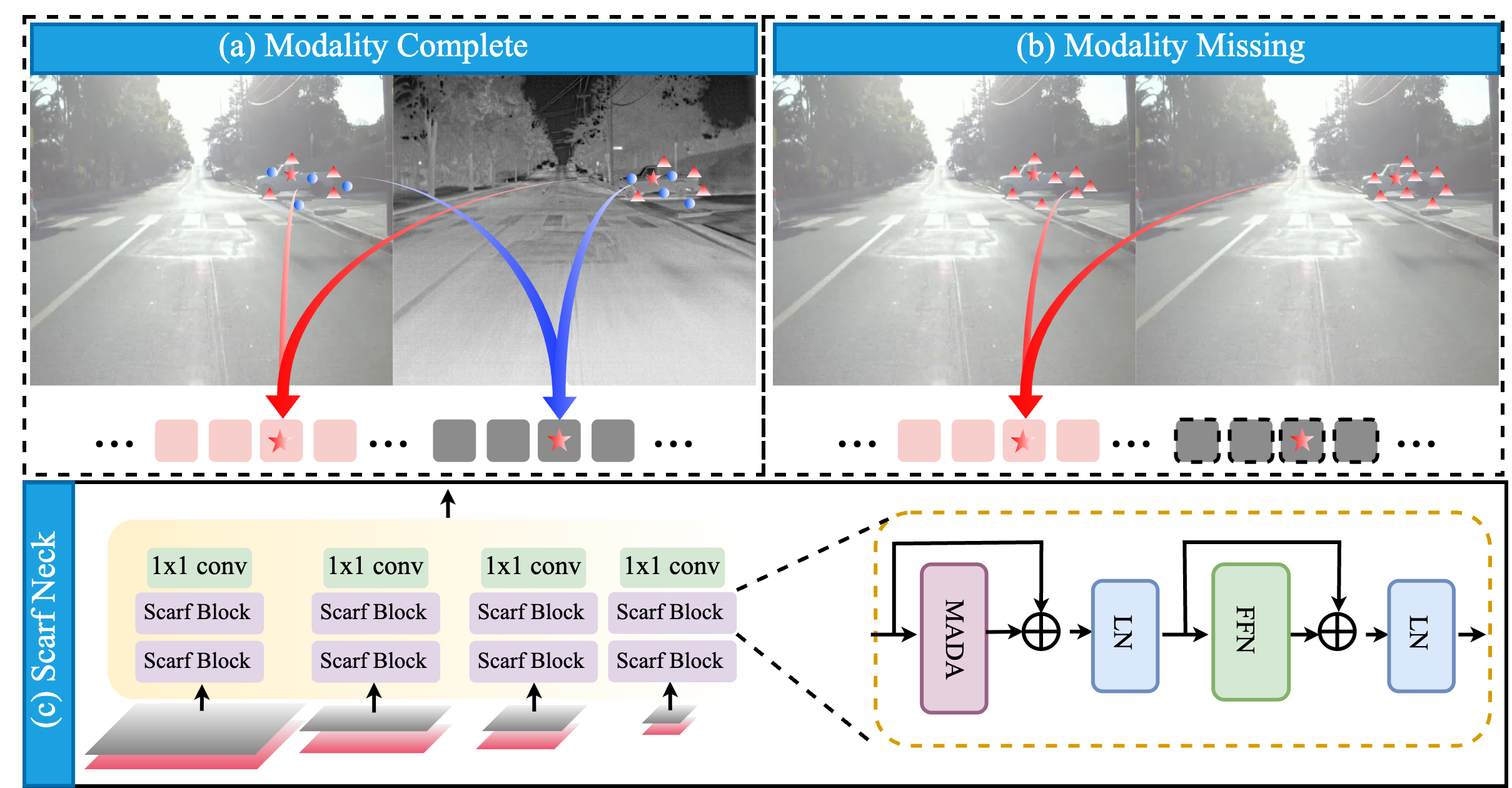}
   \caption{Overview of Scarf Neck. (a) In the complete-modality scenario, the Scarf Neck updates features of both modalities by considering intra-modal enhancement and inter-modal feature interaction.(b) In the modality-missing scenario, the Scarf Neck focuses on intra-modal feature enhancement. (c) Overview of our plug-and-play Scarf Neck.}
   \label{fig:overview}
\end{figure*}

\section{Related Work}

\noindent \textbf{Infrared-Visible Object Detection.}
Existing IVOD methods primarily focus on designing appropriate fusion strategies and loss functions to obtain better feature representations.
Some methods utilize illumination information or predict the uncertainty to act as fusion weights for different modality~\cite{yang2022baanet,drone22,zhou2020improving}, for example, Sun et al.~\cite{drone22} used the uncertainty-aware module to merge multi-modal features as well as handle redundant information.
Chen et al.~\cite{chen2022multimodal} proposed a probability-based fusion technology to combine detection results from visible and infrared modalities.
Recent researches~\cite{xing2024ms,fu2024cf,dams24} present a point-based fusion strategy, which demonstrates robustness to misalignments between visible and infrared images.  
However, most existing methods require complicated network structures but lack the capability to address the problem of modality incompleteness, which is a common issue in real-world scenarios. 
\noindent \textbf{Multimodal Learning with Missing Modality.}
Existing methods for handling missing modalities can be broadly categorized into two aspects:
1) \emph{Data processing approaches}, which aim to compensate for missing modality information by compositing or generating absent modalities~\cite{ma2021smil,zheng2021robust}.
2) \emph{Strategy design approaches}, which focus on combining modality-specialized models or devising flexible architectures that can adapt to various modalities. 
The combination-based methods employ ensemble models trained on partial or full sets of modalities, and select the most suitable model according to the available modalities~\cite{chen2022multimodal,xue2023dynamic}.
In contrast, architecture-based approaches leverage attention mechanisms to dynamically adjust modality fusion~\cite{ge2023metabev}, or utilize knowledge distillation to transfer knowledge from full-modality models~\cite{wang2023multi}.
However, the missing modality problem remains insufficiently explored in the field of IVOD. MiPa~\cite{medeiros2025mixed} attempted to address a similar issue related to modality imbalance that bears resemblance to partial modality missing scenarios, but it exclusively focused on single-modality inference and completely overlooked the scenario where dual modalities are available for inference.
Inspired by the scalability and flexibility of architecture-based approaches, we propose a Scarf Neck module to effectively address modality missing in IVOD scenarios.
\begin{figure*}[t]
  \centering
   \includegraphics[width=1.0\linewidth]{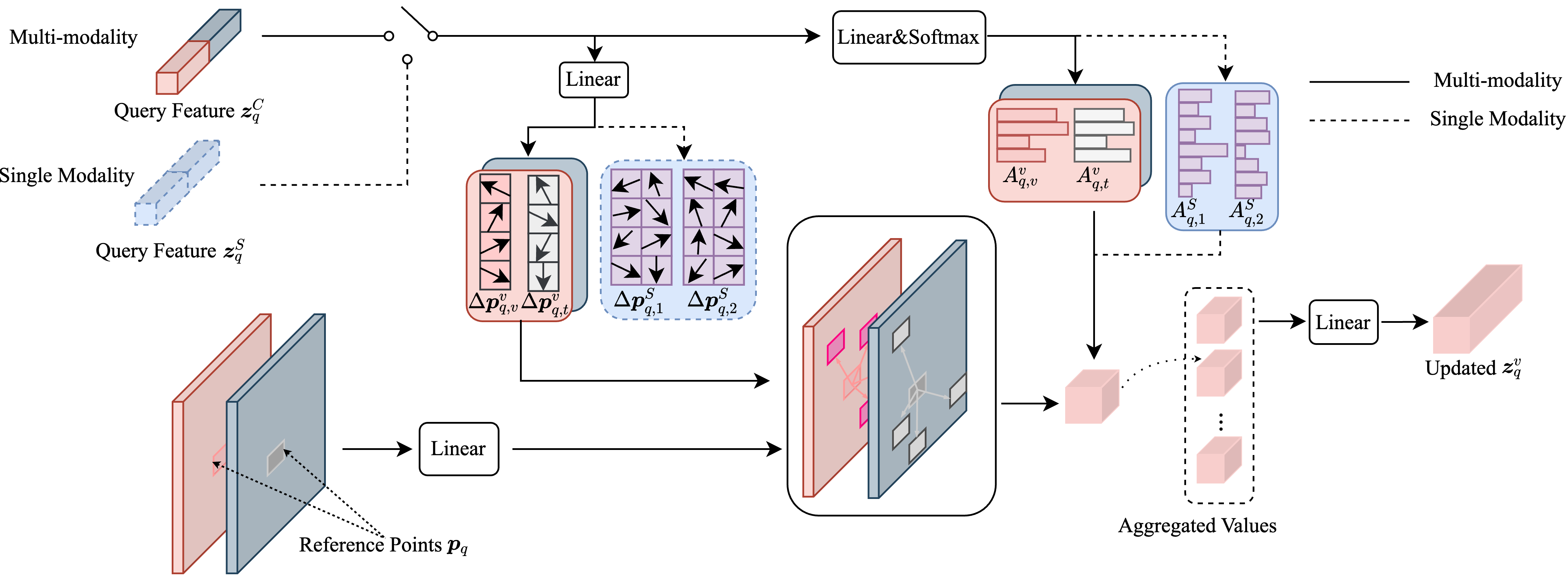}
   \caption{Details of Modality-Agnostic Deformable Attention. 
    In the multi-modality case, we obtain $\{\Delta \boldsymbol{p}^{v}_{q,\cdot}, \boldsymbol{A}^{v}_{q,\cdot}\}$ and $\{\Delta \boldsymbol{p}^{t}_{q,\cdot}, \boldsymbol{A}^{t}_{q,\cdot}\}$ to update visible features and infrared features respectively. If one modality is missing, we obtain $\{\Delta\boldsymbol{p}^S_{q,1},\boldsymbol{A}^S_{q,1}\}$ and $\{\Delta\boldsymbol{p}^S_{q,2},\boldsymbol{A}^S_{q,2}\}$ to update the available features.
   }
   \label{fig:bi_details}
\end{figure*}

\section{Scarf-DETR}
\label{sec:methods}
\subsection{DETR with Scarf Neck}  
Our model is built on DETR variants. The structure of the Scarf Neck, illustrated in Figure~\ref{fig:overview}(c), consists of multiple groups. Each group, comprising two Scarf Blocks and one 1$\times$1 convolution, processes multimodal features at a specific scale. The Scarf Block is structurally similar to the Transformer Block. 

Scarf-DETR, a DETR model integrated with the Scarf Neck, features a straightforward working mechanism: visible (VIS) and infrared (IR) images share the same backbone to extract multi-scale features for both modalities. After these features are fed into the Scarf Neck, the update of VIS features simultaneously handles intra-modal feature enhancement and inter-modal feature interaction, and the same applies to the update of IR features, as shown in Figure~\ref{fig:overview}(a). Once features at all scales are fused, the updated VIS and IR features are merged via a 1$\times$1 convolution. Finally, the features are fed into the encoder and decoder to generate detection results. 

In cases of missing modalities, Scarf-DETR focuses solely on further enhancing unimodal features without considering inter-modal interaction, as shown in Figure~\ref{fig:overview}(b): all obtained sampling points are used to update the available modality, thereby further strengthening the semantic information of the features. After the update of unimodal features, the current features are copied and concatenated along the channel dimension to adapt to the subsequent 1$\times$1 convolution.


\subsection{Modality-Agnostic Deformable Attention}
As the core operation in Scarf Neck, modality-agnostic deformable attention (MADA) is motivated from the multi-scale deformable attention mechanism. In deformable attention~\cite{deformable21}, the query feature focuses only on a small part of feature points around its reference points. 
In this paper, we improve the generalization of DETR variants by imitating the multi-scale interaction mechanism to develop a compatible modality interaction mechanism within the Scarf Neck. 
Following this, we present the concept of modality-agnostic deformable attention, considering both modality-complete and modality-incomplete scenarios.

\noindent\textbf{Modality-Complete Condition.} In modality-complete case, there are two modality features, $\boldsymbol{x}^{v}$, $\boldsymbol{x}^{t}\in \mathbb{R}^{c\times H \times W}$ after the backbone network, where $H$, $W$ and $c$ represent the height, width and dimension of the feature, respectively. ``$v$" and ``$t$" denote the visible and infrared modalities.
We flatten them into two series of tokens $\boldsymbol{E}^v = [\boldsymbol{e}^v_1,\boldsymbol{e}^v_2,\cdots, \boldsymbol{e}^v_L]^T \in \mathbb{R}^{L \times c}$ and $\boldsymbol{E}^t=[\boldsymbol{e}^t_1,\boldsymbol{e}^t_2,\cdots, \boldsymbol{e}^t_L]^T \in \mathbb{R}^{L \times c}$, where $L=H \times W$. 
Let $q \in \{1, 2, \cdots, L\}$ be the token index, we concatenate tokens of two modalities in the same spatial position $\boldsymbol{e}^{v}_{q} \in \mathbb{R}^c$ and \(\boldsymbol{e}^{t}_{q}\in \mathbb{R}^c\) to form \(\boldsymbol{z}^{C}_{q} \in \mathbb{R}^{2c}\), with the superscript $C$ denoting ``Combined" features.
Like deformable DETR~\cite{deformable21}, $\boldsymbol{z}^{C}_{q}$ passes through two linear projectors to obtain the sampling point offsets \(\Delta \boldsymbol{p}^{C}_{q} \in \mathbb{R}^{ m \times 4 \times K \times 2}\) and their corresponding weights \(\boldsymbol{A}^{C}_{q} \in \mathbb{R}^{m \times 4 \times K}\), where \(K\) represents the number of sampling points, and \(m\) is the index of the attention heads. As shown in Figure~\ref{fig:bi_details}, for each attention head, the sampling points and attention weights are split into two parts, serving for each modality: $\{\Delta \boldsymbol{p}^{v}_{q} \in \mathbb{R}^{ m \times 2 \times K \times 2}, \boldsymbol{A}^{v}_{q} \in \mathbb{R}^{ m \times 2 \times K}\}$ and $\{\Delta \boldsymbol{p}^{t}_{q} \in \mathbb{R}^{ m \times 2 \times K \times 2}, \boldsymbol{A}^{t}_{q} \in \mathbb{R}^{ m \times 2 \times K}\}$. 
Following \cite{deformable21}, we perform $\operatorname{softmax}$ function on the attention weights to obtain normalized attention weights. The attention features of visible and infrared modalities are calculated by

\begin{equation}
\begin{split}
&\operatorname{MADA}_v(\boldsymbol{z}^{C}_{q},\boldsymbol{p}_{q},\boldsymbol{x}^{v},\boldsymbol{x}^{t};v) = \\
&\sum_{m=1}^{M} \boldsymbol{W}_{m} [\sum_{S \in \{v, t\}} \sum_{k=1}^{K} A^{v}_{mqkS} \cdot \boldsymbol{W}'_{m}\boldsymbol{x}^{S}(\boldsymbol{p}_{q} + \Delta \boldsymbol{p}^{v}_{mqkS})], \\
\end{split}
\end{equation}
\begin{equation}
\begin{split}
&\operatorname{MADA}_t(\boldsymbol{z}^{C}_{q},\boldsymbol{p}_{q},\boldsymbol{x}^{v},\boldsymbol{x}^{t};t) = \\ 
&\sum_{m=1}^{M} \boldsymbol{W}_{m} [\sum_{S \in \{v, t\}} \sum_{k=1}^{K} A^{t}_{mqkS} 
\cdot \boldsymbol{W}'_{m} \boldsymbol{x}^{S}(\boldsymbol{p}_{q} + \Delta \boldsymbol{p}^{t}_{mqkS})],
\end{split}
\end{equation}
where \(\boldsymbol{p}_{q} \in \mathbb{R}^2\) is the reference point, which is the the position of the \(\boldsymbol{e}^{v}_q\) (or \(\boldsymbol{e}^{t}_q\) ) in the \(2D\) feature. \(\boldsymbol{W}_{m} \in \mathbb{R}^{c \times d_{head}}\) and \(\boldsymbol{W}'_{m} \in \mathbb{R}^{d_{head} \times c}\) are learnable weights. 



\noindent\textbf{Modality-Incomplete Condition.} 
Given a certain modality feature map from the backbone \( \boldsymbol{x}^S \in \mathbb{R}^{c \times H \times W} \), where \( S \in \{v,t\}\) is either visible or infrared modality, we first reshape it into a token sequence as $\boldsymbol{E}^S = [\boldsymbol{e}^S_1,\boldsymbol{e}^S_2,\cdots, \boldsymbol{e}^S_L]^T \in \mathbb{R}^{L \times c}$. 
In order to maintain compatibility with the modality-complete data, we stack each \( \boldsymbol{e}^S_q \) to obtain \( \boldsymbol{z}^S_q \in \mathbb{R}^{2c} \), as illustrated in Figure 3. Similarly, through linear projectors, we get \( \Delta \boldsymbol{p}_q \in \mathbb{R}^{m \times 4 \times K \times 2} \) and the corresponding weights \( \boldsymbol{A}_q \in \mathbb{R}^{m \times 4 \times K} \). They are also split into two sets of sampling points \( \Delta \boldsymbol{p}^1_q \in \mathbb{R}^{m \times 2 \times K \times 2} \) and \( \Delta \boldsymbol{p}^2_q \in \mathbb{R}^{m \times 2 \times K \times 2} \). Although only one modality is available, we can still utilize the two sets of sampling points and weights by a simple supplementation and split.
We merge the sampling points and corresponding weights into one set \{\( \Delta \boldsymbol{p}_{q}^S \in \mathbb{R}^{m \times 2 \times 2K \times 2} \), \( \boldsymbol{A}^S_q \in \mathbb{R}^{m \times 2 \times 2K} \)\}. The attention features are obtained through
\begin{equation}
\begin{split}
&\operatorname{MADA}(\boldsymbol{z}^S_q, \boldsymbol{p}_q, \boldsymbol{x}^{S,1} , \boldsymbol{x}^{S,2};S) = \\ 
&\sum_{m=1}^{M} \boldsymbol{W}_m \left[ \sum_{i=1}^{2} \sum_{k=1}^{2K} A^{S}_{mqki} \cdot \boldsymbol{W}'_m \boldsymbol{x}^{S,i}(\boldsymbol{p}_q + \Delta \boldsymbol{p}^{S}_{mqki}) \right], 
\end{split}
\end{equation}

Compared to the modality-complete case, we only need to double the sampling points regardless of which modality is missing. Through such a simple design, the MADA achieves compatibility under both modality-complete and modality-incomplete conditions, with redundant calculations minimized as much as possible.


\begin{figure}[h]
  \centering
   \includegraphics[width=0.8\linewidth]{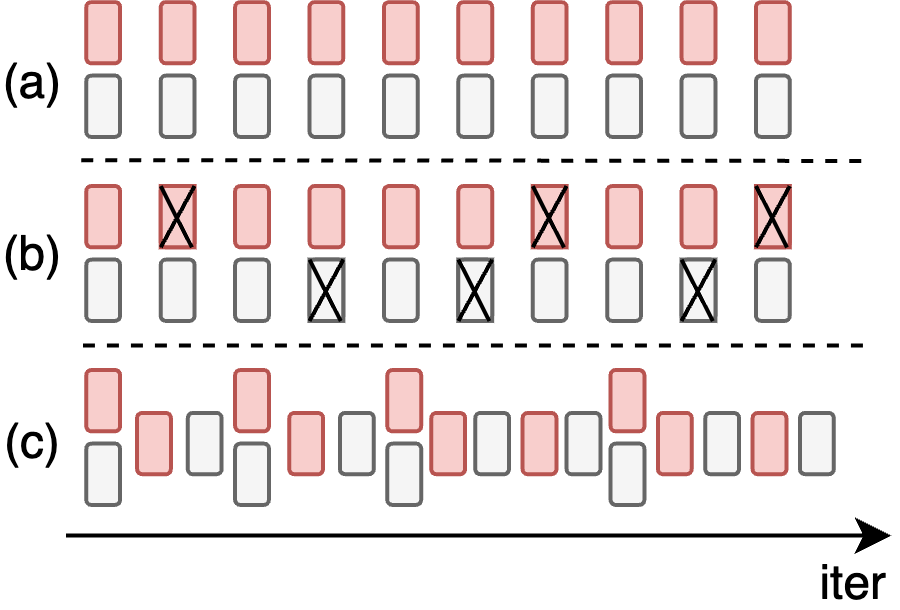}
   \caption{Illustration of joint training with (a) full modality paired data, (b) vanilla modality dropout strategy, and (c) our proposed pseudo modality dropout strategy.}
   \label{fig:problem_of_real_dp}
\end{figure}


\begin{table*}[ht]\footnotesize
\centering
\setlength{\tabcolsep}{0.5mm} 
\begin{tabular}{c  c  c  ccc  ccc  ccc  ccc  ccc } 
\hline
\multicolumn{1}{c|}{\multirow{2}{*}{\textbf{Dataset}}}  & \multicolumn{1}{c|}{\multirow{2}{*}{\textbf{Method}}} & \multicolumn{1}{c|}{\multirow{2}{*}{\textbf{Venue}}}  & \multicolumn{3}{c|}{\textbf{0\%VIS-100\%IR}} & \multicolumn{3}{c|}{\textbf{30\%VIS-70\%IR}} &  \multicolumn{3}{c|}{\textbf{50\%VIS-50\%IR}} & \multicolumn{3}{c|}{\textbf{70\%VIS-30\%IR}}  & \multicolumn{3}{c}{\textbf{100\%VIS-0\%IR}} \\ 
\cline{4-18} 
\multicolumn{1}{c|}{} &   \multicolumn{1}{c|}{}  & \multicolumn{1}{c|}{}  & mAP & 50 &\multicolumn{1}{c|}{75} & mAP & 50 &\multicolumn{1}{c|}{75} & mAP & 50 &\multicolumn{1}{c|}{75} & mAP & 50 &\multicolumn{1}{c|}{75} & mAP & 50 &75 \\
\hline
\multirow{4}{*}{\makecell[c]{FLIR-\\aligned}} 
& CFT & PR22 & 31.0 & 63.3 & 24.8 & 28.1 &60.3 &21.5  &26.7 & 58.6  &20.1   &24.6 &56.2 &17.9 & 23.6 &55.7 &16.6  \\ 
& ICAFusion &PR24  & 34.8  & 72.9 & 27.9 & 31.6 &69.2 &24.1  &29.3 & 67.3 &21.6  &27.1 &63.6 &18.9 &24.4 & 60.3 & 16.5   \\
& DAMS-DETR &ECCV24  & 47.0 & 82.6 & \textbf{45.5} & 40.2 &  77.1 &  35.6    &  35.9 &71.4 & 31.5 &  32.8 & 67.4 & 27.4  &  28.4 & 61.4  & 22.4 \\
& Scarf-Align-DETR & -   & \textbf{47.3} & \textbf{83.1} & 45.4 &  \textbf{44.6}   &  \textbf{80.8}     &  \textbf{41.2}    &  \textbf{42.8}   & \textbf{79.7} & \textbf{38.4} & \textbf{41.0} & \textbf{78.7} & \textbf{35.7}  & \textbf{37.8}  & \textbf{76.5}   & \textbf{32.3}  \\
\hline
\multirow{4}{*}{M$^3$FD} 
& CFT & PR22 & 12.6 & 23.8 & 12.4 & 17.9 & 32.5 & 17.4 & 23.9 & 31.2  & 23.8 & 27.5 & 48.4 & 26.8 & 34.5 & 59.4 & 34.4  \\ 
&ICAFusion &PR24  & 20.3 & 39.5 & 19.0 & 21.1 & 41.4 & 19.2 & 22.6 & 44.0 & 20.7 &23.3 & 45.4 & 21.5 &25.0  & 48.6 & 23.0  \\
& DAMS-DETR &  ECCV24 & 26.9  & 44.4 & 27.5 &33.8   & 53.8 & 35.2 & 36.3 &  57.4 &  37.9 & 41.5  &64.7  &43.5  & 46.6 & 71.6 & 48.8 \\
& Scarf-Align-DETR  & - & \textbf{41.0} & \textbf{65.2} & \textbf{42.0} &\textbf{43.9}   & \textbf{68.4} & \textbf{45.4} & \textbf{46.2} &  \textbf{70.8} &  \textbf{48.6} & \textbf{48.3}  &\textbf{73.6}  &\textbf{50.9} &\textbf{51.4} & \textbf{77.5} & \textbf{54.9} \\
\hline
\multirow{4}{*}{LLVIP}  
& CFT       &PR22    & 60.4 & 94.0 & 66.2 &50.1  &82.1 &53.2 &42.6 &73.5 &43.0  &34.5  &63.8  &32.6 & 21.8 & 49.0 & 16.2  \\
& ICAFusion &PR24    & 63.4 & 97.2 & 73.1 & 44.3 & 68.9 & 50.6 & 32.5 & 51.5 & 36.7 & 19.7 & 32.1 & 22.2 &1.0 &2.7 &0.0    \\
& DAMS-DETR &ECCV24  & \textbf{69.3} & 97.6 & 78.3 &54.0  &82.8 &57.8 &42.8 &69.9 &43.7  &33.1  &59.8  &31.0 & 15.0 &38.1 &8.7    \\
& Scarf-Align-DETR & -   & \textbf{69.3} & \textbf{98.2} & \textbf{82.2}  &\textbf{65.9}  &\textbf{96.8} &\textbf{76.4} &\textbf{63.3} &\textbf{95.7} & \textbf{72.2} &\textbf{60.2}  &\textbf{94.5}  &\textbf{67.5}  &\textbf{55.5} &\textbf{92.6} &\textbf{60.3}  \\                     
\hline
\end{tabular}
\caption{The results on all Modality Incomplete (-MI) benchmarks. Best results are highlighted in bold.} 
\label{tab:benchmark}
\end{table*}

\subsection{Pseudo Modality Dropout}
Most IVOD models are trained on well-aligned image pairs, as shown in Figure~\ref{fig:problem_of_real_dp}(a), but such models tend to exhibit significant vulnerability in modality-missing scenarios. The vanilla modality dropout strategy illustrated in Figure~\ref{fig:problem_of_real_dp}(b) can effectively alleviate this issue: dual-branch models can implement it by feeding samples of the same modality or using random zero-padding, while models with a shared backbone can directly remove one of the paired images to improve computational efficiency. However, vanilla modality dropout leads to a loss of sample diversity—with a 60\% dropout rate, up to 30\% of images are excluded from training. To address this inefficiency, we propose a pseudo modality dropout strategy. Specifically,
in the absence of modality dropout, each image pair is concatenated along the channel dimension to form \(\mathbf{I}^{C} = \text{concate}(\mathbf{I}^{v}, \mathbf{I}^{t}) \in \mathbb{R}^{bs \times 6 \times H \times W}\), with $bs$ denoting the batch size. If modality dropout occurs, we merely disregard the modality relationship, separating the pairs into individual samples for training, thereby creating $[\mathbf{I}^{v}; \mathbf{I}^{t}] \in \mathbb{R}^{2bs \times 3 \times H \times W}$, sharing the same ground truth label, as illustrated in Figure~\ref{fig:problem_of_real_dp}(c).

\section{Experiments}
\label{sec:exp}
\noindent\textbf{Datasets.}
We use three datasets: two variants of the FLIR dataset (FLIR~\cite{re22} and FLIR-aligned~\cite{zhang2020multispectral}), M\(^3\)FD~\cite{liu2022target} with dataset splits following~\cite{dams24}, and LLVIP~\cite{llvip}. 
These datasets are well-aligned and widely adopted in related tasks, ensuring robustness in method evaluation. Detailed information about these datasets can be found in the Appendix.

\noindent\textbf{Evaluation Metrics.} %
We mainly report mAP as the detection metric, with IoU thresholds ranging from 0.5 to 0.95. Additionally, to verify the model's robustness in modality-incomplete case, we conduct extensive experiments under three conditions: Complete modality, VIS-only, and IR-only. In order to further simulate the real-world scenarios of missing modalities, we have established the MI benchmark to evaluate the model's robustness in real-world scenarios more comprehensively.
\subsection{Implementation Details}
Our approach builds upon the MMDetection toolbox~\cite{mmdetection}. We incorporate the Scarf Neck into mainstream DETR models, specifically Align-DETR~\cite{align} and \(\mathcal{C}\)o-DETR~\cite{codetr}, to verify its effectiveness.
For key configurations, we employ 2 Scarf Blocks at each scale, with the number of sampling points K=4. 
Models are trained for 12 epochs with AdamW (learning rate\(10^{-4}\)) and standard scale jitter for data augmentation. 
Unless otherwise specified, the default backbone is Swin-S~\cite{swin} pre-trained on ImageNet-1k.



\subsection{On Modality Incomplete scenarios}
\noindent\textbf{VIS/IR-only Scenarios} The VIS/IR-only scenarios, corresponding to the ``100\%VIS-0\%IR" and ``0\%VIS-100\%IR" settings in Table~\ref{tab:benchmark}, evaluate model performance when inputs are restricted to a single modality.

On FLIR-aligned, Scarf-Align-DETR exhibits comparable performance to DAMS-DETR in the IR-only setting. However, in the VIS-only scenario, DAMS-DETR’s performance drops sharply to 28.4\%, revealing a notable over-reliance on the IR modality. A similar pattern of modality bias is observable in CFT and ICAFusion. In contrast, Scarf-Align-DETR mitigates this single-modality dependency, achieving a VIS-only mAP of 37.8\%. Here, the primary challenge shifts to the inherent information limitation of the VIS modality itself, rather than the model’s over-reliance on a specific modality, a critical distinction that underscores its robustness.


Notably, the trend reverses for M$^3$FD: here, the VIS modality plays a more critical role in detection. This can be attributed to the dataset’s target categories—such as traffic lights—whose visual characteristics are inherently more dependent on visible spectrum information. Methods like DAMS-DETR exhibit strong VIS-dominance, leading to suboptimal handling of IR-modal inputs: in the IR-only scenario, DAMS-DETR achieves only 26.9\% mAP. In contrast, our Scarf-Aligned-DETR maintains a robust 41.0\% mAP in IR-only settings. 

For LLVIP, other methods exhibit a severe IR-dominance, which we attribute to the fact that the IR alone often provides sufficient information for pedestrian detection—leading models to neglect the processing of VIS inputs. Specifically, while DAMS-DETR achieves IR-only performance close to its modality complete performance, its VIS-only mAP is even lower than that of CFT, highlighting a critical fragility in handling VIS-dependent scenarios. In contrast, our Scarf-Aligned-DETR maintains a strong 55.5\% mAP in the VIS-only setting, effectively addressing the underemphasis on VIS processing—a key improvement that underscores its balanced modality adaptation.




\noindent\textbf{MI benchmark.}
The performance metrics of VIS/IR-only scenarios are critical for evaluating model robustness, but to further simulate real-world modality-missing scenarios—where modality incompleteness is often partial rather than absolute—we additionally introduce three mixed-modality settings: 30\%VIS+70\%IR, 50\%VIS+50\%IR, and 70\%VIS+30\%IR, as a complement to VIS/IR-only evaluations. We tested existing methods across these settings on three datasets, yielding the FLIR-MI, M$^3$FD-MI, and LLVIP-MI benchmarks as shown in Table~\ref{tab:benchmark}.
A notable trend emerges: model performance generally follows a linear combination of the performances of the two single modalities (i.e., VIS-only and IR-only), weighted by their respective proportions in the test data.
Our method exhibits smaller performance fluctuations across all mixed settings. For instance, on FLIR-MI, while DAMS-DETR’s mAP drops by around 7.4\% when shifting from 70\%IR to 70\%VIS, our method maintains a more stable performance with a drop of only 3.6\%. Analogous trends hold for the other datasets.

\subsection{On Modality Complete scenarios}
In this section, Scarf-DETR is compared with the state-of-the art methods on three public benchmarks in the multi-modality setting. These advanced methods are generally designed specifically for scenarios without modality missing. Therefore, we only compare the performance of Scarf-DETR with these methods under the modality-complete test setting.

\noindent\textbf{FLIR.}
For the FLIR dataset, our method is compared with CFR~\cite{zhang2020multispectral}, GAFF~\cite{zhang2021guided}, RGBT Network~\cite{re22}, and EME~\cite{rezx24}. As shown in Table~\ref{tab:flir}, Scarf-Align-DETR outperforms RGBT Network~\cite{re22} by 2.8\% in mAP50. Additionally, under COCO pre-training, our method surpasses EME~\cite{rezx24} by 5.2\% in mAP50. Consequently, Scarf-\(\mathcal{C}\)o-DETR achieves an mAP of 56.2\%.
For the FLIR-aligned dataset, we compare our method with CF-Deformable DETR~\cite{fu2024cf}, MFPT~\cite{zhu2023multi}, LRAF-Net~\cite{fu2023lraf}, CFT~\cite{qingyun2021cross}, TFDet~\cite{zhang2024tfdet}, ICAFusion~\cite{shen2024icafusion}, FD\(^2\)Net~\cite{li2024fd2}, and DAMS-DETR~\cite{dams24}. 
We observe that Scarf-Align-DETR demonstrates performance comparable to LRAF-Net~\cite{fu2023lraf}. Furthermore, under COCO pre-training, our method matches the performance of DAMS-DETR~\cite{dams24}. Both achieve this while maintaining robustness to missing modality. Accordingly, Scarf-\(\mathcal{C}\)o-DETR yields an mAP of 51.2\%.
\begin{table}[t]\footnotesize
\centering
\setlength{\tabcolsep}{0.2mm} 
\begin{tabular}{ c | c | c c c | c c c} 
\hline
\multirow{2}{*}{\textbf{Dataset}} &\multirow{2}{*}{\textbf{Methods}} & \multirow{2}{*}{\textbf{mAP}} 
&\multirow{2}{*}{\textbf{50}} & \multirow{2}{*}{\textbf{75}}
&\multicolumn{3}{c}{\textbf{AP50}} \\ 
\cline{6-8}
 &  &    &     &  &Bicycle & Car & Person \\ 
\hline
\multirow{7}{*}{FLIR} 
& CFR & - &72.3 &- &57.7  &84.9 &74.4\\ 
&GAFF & - &72.9 &- &-       &-    &-  \\ 
&RGBT Network & - &82.9 &- &76.7    &91.8 &80.1\\ 
&Scarf-Align-DETR & \textbf{50.1} &\textbf{85.7} &\textbf{50.4} &\textbf{79.1}  &\textbf{92.3} &\textbf{85.7}\\ 
\cline{2-8}
&EME\(^{*}\) & - &84.6 &- &79.7  &92.6 &81.2\\ 
&Scarf-Align-DETR\(^{*}\) & \textbf{55.4} &\textbf{89.8} &\textbf{57.4} &\textbf{84.8}  &\textbf{94.6} &\textbf{89.9}\\
\cline{2-8}
&Scarf-$\mathcal{C}$o-DETR\(^{\dagger}\) & 56.2 &90.6 &58.7 &86.2  &94.8 &90.6\\
\hline
\hline
\multirow{10}{*}{\makecell[c]{FLIR-\\aligned}}
&\makecell{CF-Deform. DETR}  & - &77.2 &- &-  &- &-\\ 
&MFPT  & - &80.0 &- &67.7  &89.0  &83.2  \\ 
&LRAF-Net  & 42.8 &80.5 &- &-  &- &- \\ 
&Scarf-Align-DETR & \textbf{42.9} &\textbf{80.9} &\textbf{38.3} &\textbf{69.0}  &\textbf{90.1} &\textbf{83.5}\\ 
\cline{2-8}
&CFT\(^{*}\)  & 40.2 & 78.7 & 35.5 &-  &- &-\\
&TFDet\(^{*}\)  & -   &81.7 &41.3 &71.9  &87.5 &85.2\\
&ICAFusion\(^{*}\) & - &82.8 &- & 73.8 &89.8 &84.9\\ 
&FD\(^2\)Net\(^{*}\)  &- &82.9 &42.5 &73.2  &89.9 &85.3\\
&DAMS-DETR\(^{*}\) &49.3 &\textbf{86.6} &48.1 &-  &- &-\\
&Scarf-Align-DETR\(^{*}\) & \textbf{49.6} &85.4 &\textbf{48.9} &\textbf{73.8}  &\textbf{92.8} &\textbf{89.5}\\ 
\cline{2-8}
&Scarf-$\mathcal{C}$o-DETR\(^{\dagger}\) & 51.2 & 87.9 &50.2 &78.6  &93.7 &91.5\\
\hline
\end{tabular}
\caption{Comparisons on FLIR dataset. Models with * denote COCO pre-trained; $\dagger$ refers to Object365\&COCO pre-trained. Best results are highlighted in bold.}
\label{tab:flir}  
\end{table}

\noindent\textbf{M\(^3\)FD.}
The results on M\(^3\)FD dataset is shown in Table~\ref{tab:m3fd}. 
We follow the dataset split from~\cite{dams24}, with comparison methods including ICAFusion~\cite{shen2024icafusion}, CFT~\cite{qingyun2021cross}, and DAMS-DETR~\cite{dams24}. 
Scarf-Align-DETR, initialized with ImageNet-1k pre-trained weights, notably exhibits a 1.5\% mAP advantage over CFT~\cite{qingyun2021cross}, which uses COCO pre-trained weights. Conversely, under the consistent COCO pre-training setting, our method outperforms DAMS-DETR~\cite{dams24} by 1.5\% in mAP. Our best model attains a state-of-the-art mAP score of 57.1\%. 
\begin{table}[t]\footnotesize
\centering
\setlength{\tabcolsep}{1.0mm} 
\begin{tabular}{ c | c c c } 
\hline
\textbf{Methods} & \textbf{mAP}  & \textbf{mAP50} & \textbf{mAP75} \\ 
\hline
ICAFusion* & 41.9 &67.8 &44.5 \\ 
CFT*      & 42.5 &68.2 &44.6 \\ 
Scarf-Align-DETR & 44.0 &70.5 &45.5\\ 
DAMS-DETR\(^{*}\)  &52.9 &80.2 & 56.0 \\ 
Scarf-Align-DETR\(^{*}\) & \textbf{54.4} &\textbf{80.9} &\textbf{57.5}\\
\cline{1-4}
Scarf-$\mathcal{C}$o-DETR\(^{\dagger}\) & 57.1 &83.8 &61.7 \\  
\hline
\end{tabular}
\caption{Comparisons on  M\(^3\)FD dataset. Models with * denote COCO pre-trained; $\dagger$ refers to Object365\&COCO pre-trained. Best results are highlighted in bold.}
\label{tab:m3fd}  
\end{table}

\noindent\textbf{LLVIP.}
 We provide the results on LLVIP datasets in Table~\ref{tab:llvip}. Compared with CSAA~\cite{cao2023multimodal}, CF-Deformable DETR~\cite{fu2024cf}, LRAF-Net~\cite{fu2023lraf}, ICAFusion~\cite{shen2024icafusion}, CFT~\cite{qingyun2021cross}, FD\(^2\)Net~\cite{li2024fd2}, MS-DETR~\cite{xing2024ms}, and DAMS-DETR~\cite{dams24}. Scarf-Align-DETR outperforms LRAF-Net~\cite{fu2023lraf} by 1.6\% AP. Under COCO pre-training, our method further surpasses DAMS-DETR~\cite{dams24} by 0.5\% in AP50 and 3.8\% in AP75. Notably, the proposed Scarf-\(\mathcal{C}\)o-DETR achieves an AP of 70.8\%.

\begin{table}[ht]\footnotesize
\centering
\setlength{\tabcolsep}{1.0mm} 
\begin{tabular}{ c | c c c } 
\hline
\textbf{Methods} & \textbf{AP}  & \textbf{AP50} & \textbf{AP75} \\ 
\hline
CSAA            & 59.2 &94.3 &66.6 \\ 
CF-Deform. DETR & -    &95.0 &- \\ 
LRAF-Net        & 66.3 &\textbf{97.9} &- \\ 
Scarf-Align-DETR & \textbf{67.9} &96.9 &\textbf{79.8}\\ 
\hline
ICAFusion\(^{*}\)    & 63.4 &97.7 &73.0 \\ 
CFT\(^{*}\)          & 63.6 &97.5 &72.9 \\ 
FD\(^2\)Net\(^{*}\)  & -    &96.2 &70.0\\
MS-DETR\(^{*}\)      & 66.1 &97.9 &76.3\\
DAMS-DETR\(^{*}\)    & \textbf{69.6} &97.9 &79.1\\
Scarf-Align-DETR\(^{*}\) & \textbf{69.6} &\textbf{98.4} &\textbf{82.9}\\
\hline
Scarf-$\mathcal{C}$o-DETR\(^{\dagger}\) & 70.8 &97.9 &81.6 \\  
\hline
\end{tabular}
\caption{Comparisons on LLVIP dataset. Models with * denote COCO pre-trained; $\dagger$ refers to Object365\&COCO pre-trained. Best results are highlighted in bold.}
\label{tab:llvip}  
\end{table}

\begin{table}[ht]\footnotesize
\centering
\setlength{\tabcolsep}{1.2mm} 
\begin{tabular}{c|ccc|ccc|ccc}
\hline
\multirow{2}{*}{\textbf{Blocks}} 
& \multicolumn{3}{c|}{\textbf{Complete}} 
& \multicolumn{3}{|c|}{\textbf{VIS}} 
& \multicolumn{3}{c}{\textbf{IR}} \\
\cline{2-10}
& {mAP} & {50} & {75} 
& {mAP} & {50} & {75} 
& {mAP} & {50} & {75}  \\
\hline
1 & 49.5 & 85.0 & 49.0 & 36.9 & 73.9 & 31.7 & 48.3 & 83.5 & 48.3 \\
2 & 50.1 & 85.7 & \textbf{50.4} & \textbf{37.5} & \textbf{74.1} & \textbf{33.0} & 48.7 & 83.8 & 48.4 \\
4 & \textbf{50.3} & \textbf{86.3} & 50.1 & 37.1 & 74.0 & 31.9 & \textbf{49.1} & \textbf{84.7} & \textbf{49.0} \\
\hline
\end{tabular}
\caption{Investigation of different number of Scarf Blocks. Best results are highlighted in bold.}
\label{tab:number_of_blocks}
\end{table}

\subsection{Ablation Studies}

\noindent\textbf{Number of Scarf Blocks.}
We conduct ablation experiments to explore the impact of different number of Scarf blocks on the FLIR dataset.
Table~\ref{tab:number_of_blocks} provides the results, where we can find that stacking Scarf blocks can bring performance improvements. 
Considering the effectiveness and efficiency of the Scarf Neck, we stack two blocks for each scale in our default setting.


\noindent\textbf{Effectiveness of Pseudo Modality Dropout.}  
We investigate the efficacy of the proposed pseudo modality dropout strategy by comparing it with vanilla modality dropout. Table~\ref{tab:pseudo} shows the performance of Scarf-Align-DETR on the FLIR dataset with varying dropout ratios. It is evident that without modality dropout, the model shows susceptibility to modality-incomplete case, meaning the model trained on multiple modality data suffers significant performance degradation when applied to single-modality data. Introducing modality dropout mitigates these performance losses across different dropout ratios (20\%, 40\%, 60\%, 80\%, and 100\%), enhancing the model's robustness. Our pseudo modality dropout improves mAP from 0.2 to 2.0 in modality-complete case and from 0.6 to 4.0 in modality-incomplete case. Accordingly, we select the 60\% dropout ratio as our default setting, as it balances performance between modality-complete and incomplete modality scenarios.

\begin{table}[t]\footnotesize
\centering
\setlength{\tabcolsep}{0.9mm} 
\begin{tabular}{c | c | cc | cc | cc} 
\hline
\multirow{2}{*}{\textbf{Dropout ratio}}  & \multirow{2}{*}{\textbf{Pseudo}}  & \multicolumn{6}{c}{\textbf{mAP} \(\uparrow\)} \\ 
\cline{3-8}
& & Complete & \color{blue}{Gain} & VIS & \color{blue}{Gain} & IR & \color{blue}{Gain}\\  
\hline

0\% &  - &49.9 &\color{blue}{-} & 18.5 & \color{blue}{-} & 42.2 &\color{blue}{-} \\  
\hline

\multirow{2}{*}{20\%}
&  \XSolidBrush  & 50.2 &\color{blue}{-}  & 30.0 &\color{blue}{-} & 47.3 &\color{blue}{-}\\   
\cline{2-8}
& \Checkmark & 50.4 &\color{blue}{+0.2}  & 32.9 & \color{blue}{+2.9}   & 47.9 &\color{blue}{+0.6}\\
\hline
\multirow{2}{*}{40\%}
&   \XSolidBrush   & 49.4 &\color{blue}{-}  & 32.1 &\color{blue}{-} & 47.4 &\color{blue}{-} \\
\cline{2-8}
& \Checkmark & 50.1 &\color{blue}{+0.7}  & 35.4 & \color{blue}{+3.3}  & 48.4 &\color{blue}{+1.0}\\
\hline
\multirow{2}{*}{60\%}
&   \XSolidBrush   & 49.0 &\color{blue}{-}  & 34.2 &\color{blue}{-} & 47.3 &\color{blue}{-}  \\
\cline{2-8}
& \Checkmark & 50.1 &\color{blue}{+1.1} & 37.5 &\color{blue}{+3.3} & 48.7 &\color{blue}{+1.4}\\
\hline
\multirow{2}{*}{80\%}
&   \XSolidBrush  & 48.3 &\color{blue}{-}  & 34.8 &\color{blue}{-} & 46.5 &\color{blue}{-}  \\
\cline{2-8}
& \Checkmark & 50.0 &\color{blue}{+0.7}  & 38.8 & \color{blue}{+4.0} & 49.1 &\color{blue}{+2.6}  \\

\hline
\multirow{2}{*}{100\%}
& \XSolidBrush  & 44.0 &\color{blue}{-} & 36.5 &\color{blue}{-}& 47.1 &\color{blue}{-}\\
\cline{2-8}
& \Checkmark & 46.0 &\color{blue}{+2.0} & 39.8 &\color{blue}{+3.3} & 49.0 &\color{blue}{+1.9}   \\
\hline
\end{tabular}
\caption{Comparison of performance between pseudo modality dropout and vanilla modality dropout.}  
\label{tab:pseudo}
\end{table}

\noindent\textbf{Double Sampling in Modality-Incomplete Case.}
To confirm compatibility in modality-incomplete case, we employ the double sampling strategy, which involves sampling double number of points from the single-modality data.
We conduct experiments on the FLIR dataset using Scarf-Align-DETR, Scarf-\(\mathcal{C}\)o-DETR and Scarf-DDQ. Table ~\ref{tab:double_sample} shows the performances of these models with and without the strategy. It can be observed that the results of single modality are generally improved. 
Even though these improvements are marginal, utilizing such a sampling strategy is crucial for compatibility design.



\begin{table}[t]  
\centering
\setlength{\tabcolsep}{0.8mm} 

\begin{tabular}{c|c|ccc}
\hline
\multirow{2}{*}{\textbf{Model}} & \multirow{2}{*}{\textbf{Double Points}} & \multicolumn{3}{c}{\textbf{mAP}\(\uparrow\)} \\
\cline{3-5}
& & Complete & VIS & IR \\ 
\hline
\multirow{2}{*}{Scarf-Align-DETR} 
&    \XSolidBrush        & 49.8      & 37.3        & 48.5       \\
\cline{2-5}
& \Checkmark & 50.1 & 37.5 & 48.7 \\
\cline{1-5} 
\multirow{2}{*}{Scarf-\(\mathcal{C}\)o-DETR}
&     \XSolidBrush    & 49.4        & 37.2        & 48.1       \\
\cline{2-5}
& \Checkmark & 49.7 & 37.4 & 48.4 \\
\cline{1-5} 
\multirow{2}{*}{Scarf-DDQ}
&    \XSolidBrush  & 49.3       & 36.7 & 47.9 \\
\cline{2-5}
& \Checkmark & 49.4 & 36.7 & 48.2  \\
\hline
\end{tabular}
\caption{Ablations of double sampling strategy.} 
\label{tab:double_sample}
\end{table}

\section{Conclusion}
\label{sec:conclusion}
In this paper, we investigate the impact of modality missing in IVOD detectors, and propose a simple yet effective module Scarf Neck, which is composed of a modality-agnostic deformable attention designed to replace the neck of DETR variants for seamless compatibility. Combined with our pseudo modality dropout strategy to enhance training efficiency by disconnecting image pairs without losing any image samples.
Combined with the Scarf Neck,  Scarf-DETR achieves performance on par with existing IVOD models, while bringing significant improvements in modality-incomplete scenarios—consistently validated by our proposed MI benchmark, which simulates real-world partial modality missing to rigorously assess such robustness.
Note that such a flexible neck enables single-modality detectors to adjust to multi-modality detection tasks, bringing greater potential for IVOD task with the development of single-modality detectors.
Future work will explore extending our method to other multi-modal tasks (e.g., RGB-D, RGB-SAR), where its modality-agnostic design is expected to adapt effectively to diverse modality-missing scenarios. We also plan to extend the MI benchmark to domains like RGB-D and beyond, to support broader multi-modal robustness evaluation.

\bibliography{aaai2026.bib}

\clearpage
\appendix 
\section{Details on Datasets}

\textbf{FLIR.} We use the FLIR dataset aligned by Vadidar et al.
It comprises 7,381 image pairs for training, 1,056 for validation, and 2,111 for testing. Each image has a resolution of 640 \(\times\) 512 and includes the ``Person'', ``Bicycle'', and ``Car'' categories. 
Furthermore, to ensure a fair comparison with advanced methods, we also present the results using an alternative version of this dataset known as FLIR-aligned. This version maintains the same resolution and categories but features a slightly smaller dataset size, comprising 4,129 image pairs for training and 1,013 image pairs for testing.
To distinguish the first-mentioned version, the later-mentioned version is named FLIR-aligned.

\noindent\textbf{M\(^3\)FD.} The M\(^3\)FD dataset comprises 4,200 infrared-visible image pairs. Since no public split is available for this dataset, we adopt a rigorous splitting strategy following Guo et al. This split uses 3,368 pairs for training and 831 for testing. 
 The dataset contains six categories, namely ``Person", ``Car", ``Bus", ``Motorcycle", ``Traffic Light", and ``Truck".

\noindent\textbf{LLVIP.} 
The LLVIP dataset is commonly employed for tasks like image fusion and detecting pedestrians. Images are captured in surveillance settings with low lighting conditions. Each image has a resolution of 1280 \(\times\) 1024. The dataset consists of a single pedestrian category, providing 12,025 image pairs for training and 3,463 pairs for testing.


\section{More Details of Experimental Setting}

The number of attention heads in each Scarf Block aligns with the backbone across scales, with the respective head numbers for the four scales being [3, 6, 12, 24]. Additionally, due to the randomness involved in the modality dropout strategy, we fixed the random seed to ensure experimental fairness. 
For the models that use the pre-trained weights from COCO or Object365 + COCO, their backbone model is Swin-L. The data augmentation used is consistent with those adopted in DETR. In the Scarf Block, the quantity of attention heads corresponds to the feature dimension; therefore, the number of attention heads is configured as [6, 12, 24, 48].

\section{More Results}
\noindent\textbf{Compatibility with Mainstream DETR Variants and Preservation of Single-Modality Efficacy.} Leveraging its inherent transferability, the Scarf Neck integrates smoothly into mainstream DETR architectures through a straightforward neck-module replacement. Table \ref{tab:4models_4datasets} reports performance metrics of our Scarf Neck when integrated with four mainstream DETR variants across four datasets, under both modality-complete (Comp.) and modality-incomplete (VIS-only, IR-only) scenarios. Notably, in modality-incomplete settings, Scarf-DETR models retain single-modality performance comparable to their original counterparts (trained solely on single-modality data). For example, DINO achieves 35.5 mAP on FLIR under VIS-only conditions, while Scarf-DINO yields a similar 35.3 mAP; DDQ obtains 37.7 mAP on FLIR under VIS-only settings, with Scarf-DDQ maintaining 36.7 mAP. This consistency holds across datasets and scenarios, confirming that our approach preserves base models' single-modality efficacy while enhancing their multi-modal capabilities.

\section{Visualizations}
To obtain a clearer understanding of how the model performs when either modality is missing, we present visualizations of three representative scenarios drawn from the test sets of the FLIR, M$^3$FD, and LLVIP datasets. The model we used is Scarf-Align-DETR. 

For each scenario, the first row is the results without dropout strategy, and the second row is the model trained with 60\% pseudo modaltiy dropout ratio. The boxes with different colors represent different categories. 

The first set of scenarios is from the FLIR dataset, where significant missed detections can be observed in the first row under the VIS-only condition. 

The second set of scenarios is from the M$^3$FD dataset, where under VIS-only conditions, the detection boxes of two cars are severely misaligned; additionally, one car target remains undetected under VIS-only, and another under IR-only.

The third set of scenarios is from the LLVIP dataset, where the model without dropout fails to detect four pedestrians, all of which are relatively easy to detect. 

Our proposed method, incorporating pseudo modality dropout, demonstrates greater robustness, achieving consistently favorable detection results under all three conditions.

\begin{figure*}[t]
  \centering
   \includegraphics[width=1.0\linewidth]{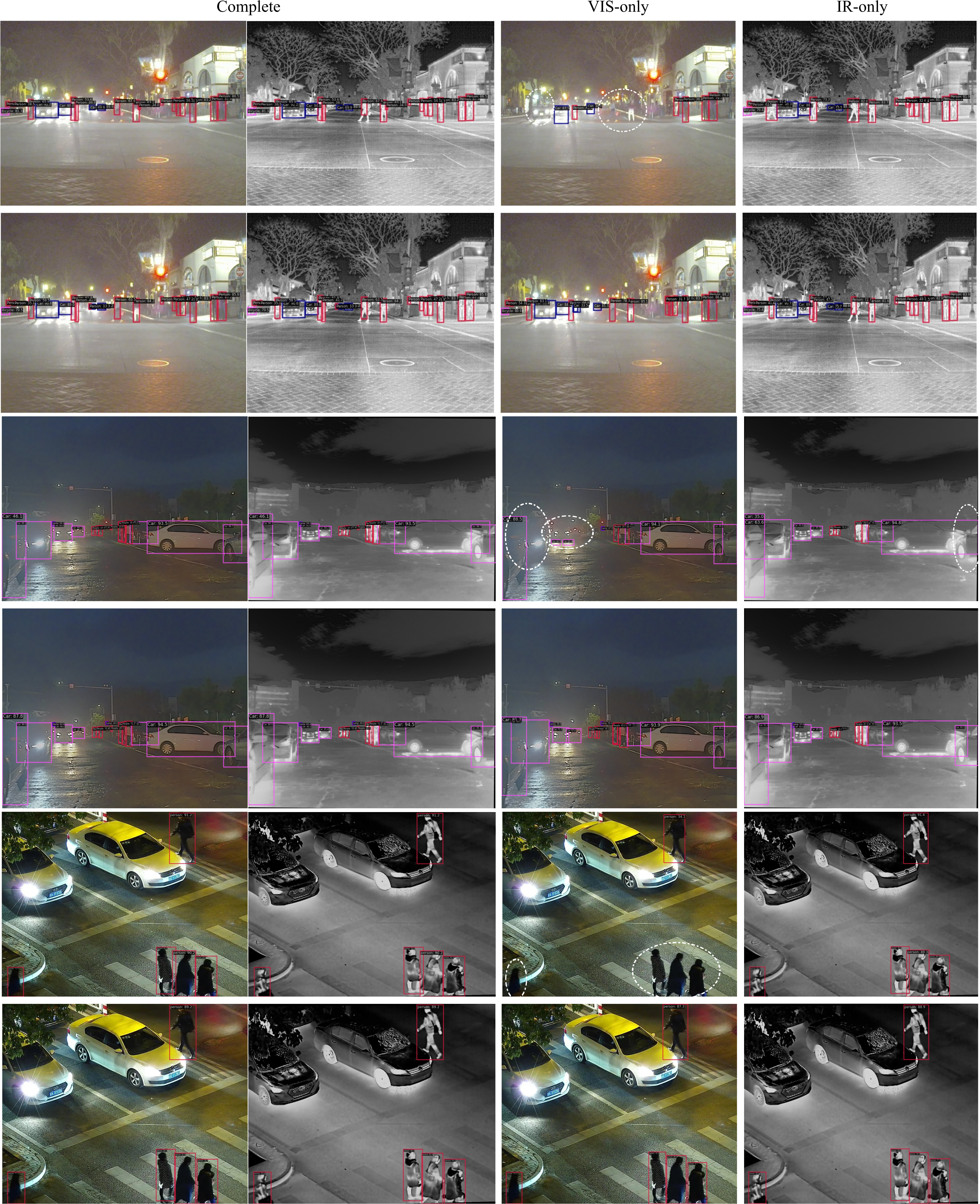}
   \caption{Visualization comparison in three scenarios. In each case, the first row is the visualizations of the model trained without dropout, and the second row is the visualizations of the model with a 60\% pseudo dropout ratio. The Scarf Neck combined with pseudo modality dropout strategy greatly improves the model's performance. Best viewed in color and with zoom for clarity.}
   \label{fig:tot_vis}
\end{figure*}


 




\begin{table}[t]\footnotesize
\centering
\setlength{\tabcolsep}{1.3mm} 
\begin{tabular}{ c| c | c| c | c | c} 
\hline
\multirow{3}{*}{\textbf{Model}}&\multirow{3}{*}{\textbf{\makecell[c]{Modal}}}  & \multicolumn{4}{c}{\textbf{mAP} \(\uparrow\)} \\ 
\cline{3-6}

& & FLIR & \makecell[c]{FLIR-\\aligned}  &\makecell[c]{M\(^3\)FD} &LLVIP \\  
\hline
\multirow{2}{*}{\makecell[c]{DINO}}  
& {VIS} & 35.5 &31.3 &35.9 &50.9 \\ 
& {IR}  & 47.0 &38.1 &29.2 &62.0  \\
\hline
\multirow{3}{*}{\makecell[c]{Scarf-\\DINO}}
& Comp. & 48.1 &41.2  &41.4 &66.6 \\  
& VIS   & 35.3 &31.1  &37.8 &53.1\\ 
& IR    & 46.6 &37.4  &29.4 &64.5\\
\hline
\multirow{2}{*}{\makecell[c]{DDQ}}   
& {VIS} & 37.7 &31.2 &40.6 &51.1 \\ 
& {IR}  & 48.5 &39.1 &31.6 &62.6  \\
\hline
\multirow{3}{*}{\makecell[c]{Scarf-\\DDQ}}
& Comp. &49.4 &43.0 &45.0 &66.8 \\  
& VIS   &36.7 &32.8 &40.4 &53.0\\ 
& IR    &48.2 &39.5 &32.2 &64.9\\
\hline
\multirow{2}{*}{\makecell[c]{$\mathcal{C}$o-DETR}}  
& {VIS} & 35.7 &30.8 &38.3 &52.6 \\ 
& {IR}  & 48.2 &38.0 &30.0 &66.4  \\
\hline
\multirow{3}{*}{\makecell[c]{Scarf-\\$\mathcal{C}$o-DETR}}
& Comp. &49.7 &44.6 &44.2 &67.9 \\  
& VIS   &37.4 &33.7 &40.0 &54.7\\ 
& IR    &48.4 &42.8 &32.4 &65.4\\
\hline
\multirow{2}{*}{\makecell[c]{Align-DETR}}
& {VIS} & 38.3 &30.8 &38.4 & 52.5 \\ 
& {IR}  & 48.9 &38.1 &30.2 &64.4  \\
\hline
\multirow{3}{*}{\makecell[c]{Scarf-\\Align-DETR}}
& Comp. & 50.1 &42.9 &44.0 &67.9 \\  
& {VIS} & 37.5 &32.9 &40.0 &53.3\\ 
& {IR}  & 48.7 &39.6 &31.8 &66.1\\
\hline
\end{tabular}
\caption{Performance of Scarf-DETR with 60\% pseudo modality dropout under modality-complete and modality-incomplete conditions, with comparison to single-modality performance of single-modality-trained counterparts. All models utilize ImageNet-1k pre-trained weights.}
\label{tab:4models_4datasets} 
\end{table}

\end{document}